\documentclass[10pt,a4paper]{article}

\usepackage[utf8]{inputenc}
\usepackage[most]{tcolorbox}
\usepackage[numbers,sort&compress]{natbib}
\usepackage[T1]{fontenc}
\usepackage[margin=0.75in]{geometry}
\usepackage{graphicx}
\usepackage{xcolor}
\usepackage{colortbl}
\usepackage{url}
\usepackage{hyperref}
\usepackage{caption}
\usepackage{subcaption}
\usepackage{booktabs}
\usepackage{array}
\usepackage{multirow}
\usepackage{ragged2e}
\usepackage{algorithm}
\usepackage[noend]{algpseudocode}
\usepackage{pifont}

\renewcommand{\arraystretch}{1.2}

\hypersetup{
  colorlinks = true,
  urlcolor   = blue,
  linkcolor  = black,
  citecolor  = blue
}

\algrenewcommand\algorithmicrequire{\textbf{Require:}}
\algrenewcommand\algorithmicreturn{\textbf{return}}

\usepackage[toc]{appendix}

\title{\textbf{R-GAT: Cancer Document Classification Leveraging Graph-Based Residual Network for Scenarios with Limited Data}}
\author{
\begin{minipage}{0.92\textwidth}
\centering
Elias Hossain$^{1,*}$, Tasfia Nuzhat$^{2}$, Shamsul Masum$^{3}$, Shahram Rahimi$^{4}$ and Noorbakhsh Amiri Golilarz$^{4}$ \\[4pt]
\small $^{1}$Department of Computer Science and Engineering, Mississippi State University, Mississippi State, MS 39762, USA \\
\small $^{2}$Department of Information Technology, Universiti Tenaga Nasional, 43000 Kajang, Selangor, Malaysia \\
\small $^{3}$School of Electrical and Mechanical Engineering, University of Portsmouth, Portsmouth, PO1 3HE, UK \\
\small $^{4}$Department of Computer Science, The University of Alabama, Tuscaloosa, AL 35487, USA \\[4pt]
\small \texttt{*Corresponding author: mh3511@msstate.edu}
\end{minipage}
}

\date{}

\begin{document}
\maketitle
\vspace{-1em}

\begin{abstract}
\noindent
Accurate classification of cancer-related biomedical abstracts is critical for advancing cancer informatics and supporting decision-making in healthcare research. Yet progress in this domain is often constrained by limited availability of labeled corpora and the high computational demands of transformer-based approaches. To address these challenges, we propose a Residual Graph Attention Network (R-GAT) that integrates multi-head attention with residual connections to capture semantic and relational dependencies in biomedical texts. Evaluated on a curated dataset of 1,875 PubMed abstracts spanning thyroid, colon, lung, and generic cancer topics, R-GAT achieves stable and competitive performance (macro-F1: 0.96 ± 0.01), comparable to transformer-based models such as BioBERT and BioClinicalBERT and strong classical baselines like Logistic Regression, while requiring significantly fewer computational resources. Ablation studies confirm the importance of attention and residual connections in ensuring robustness under limited-data conditions. To support reproducibility and facilitate future research, we also release the curated dataset. Together, these contributions demonstrate the value of lightweight graph-based architectures as reliable and resource-efficient alternatives to computationally intensive transformers in biomedical NLP.
\end{abstract}

\section{Introduction}

Cancer remains a major global health challenge, with thyroid, colon, and lung cancers ranking among the most prevalent and deadly types worldwide \cite{zhai2021global,WHOColorectalCancer,Wang2019}. The scale of this challenge has driven extensive biomedical research, resulting in a rapidly expanding body of scientific publications. Notably, most of these findings are first communicated through PubMed abstracts, which provide concise yet information-rich summaries of cancer-related studies. While such abstracts offer valuable insights for clinical and scientific progress, their growing volume and complexity make manual analysis infeasible. Consequently, automated classification of cancer abstracts has become an essential task to accelerate literature mining and knowledge discovery.

In response to this need, recent advances in natural language processing (NLP) have enabled substantial progress in biomedical text mining \cite{moqbel2025mining}, offering tools for entity recognition \cite{liu2025knowledge}, relation extraction \cite{yang2025automatically}, and document classification \cite{alva2025long}. Nevertheless, two critical obstacles remain. First, large biomedical resources such as CORD-19 \cite{wang2020cord19} are broad and noisy, while electronic health records (EHRs) are incomplete and difficult to access, leaving a gap in reliable, cancer-focused corpora. While multi-cancer datasets such as the Hallmarks of Cancer corpus (HOC) \cite{DBLP:journals/bioinformatics/BakerSGAHSK16} and CORD-19 \cite{wang2020cord19} subsets exist, they are either broad in scope or not specifically balanced across thyroid, colon, and lung cancer categories.
Second, state-of-the-art deep learning models, particularly transformers, achieve high accuracy but require massive labeled datasets and substantial computational resources, which limits their use in data-constrained biomedical contexts. More importantly, this limitation highlights the need for models that can generalize effectively without relying on large-scale training corpora.


To ensure consistent evaluation, we curated a controlled subset of PubMed abstracts, offering a reproducible benchmark for cancer classification studies. At the same time, addressing the methodological gap left by data- and compute-intensive transformers requires exploring alternative architectures. One promising direction is the use of graph-based neural networks, which differ from sequence models by representing biomedical abstracts as interconnected structures that capture relational and semantic dependencies between terms. When combined with attention mechanisms, such models can highlight critical biomedical entities and their interactions, while residual connections mitigate information loss and stabilize training. Collectively, these properties position graph-based residual architectures as strong candidates for robust text classification in scenarios where labeled data are scarce and computational resources are limited.

Building on this rationale, this study investigates the following research question: \textit{Can graph-based residual architectures provide a robust and computationally efficient alternative to transformer models for cancer abstract classification under limited-data conditions?}

To address this question, we propose R-GAT, a graph-based model that integrates multi-head attention with residual connections to capture semantic dependencies more effectively. Our approach is systematically benchmarked against a wide range of traditional machine learning, deep learning, and transformer-based baselines, with ablation studies conducted to assess the contribution of each architectural component. Ultimately, the aim of this study is to demonstrate that R-GAT provides a stable and computationally efficient solution for cancer document classification in limited-data settings. Building on this objective, the key contributions are as follows:

\begin{itemize}
    \item We introduce R-GAT, a graph-based approach that leverages residual connections and multi-head attention to deliver robust performance under limited-data biomedical NLP conditions.  
    \item We design a rigorous evaluation framework, benchmarking R-GAT against traditional machine learning, deep learning, and transformer-based baselines, and conducting ablation studies to isolate the impact of its architectural components.  
    \item To ensure reproducibility and provide a controlled testbed, we make available a curated subset of $\approx$1,875 PubMed abstracts, balanced across thyroid, colon, and lung cancers, which serves as a benchmark resource for future cancer document classification studies.  
\end{itemize}

\vspace{3pt}

The remainder of this manuscript is organized as follows. First, we review related work on cancer document classification, followed by a description of the dataset and its statistical properties. We then present the proposed R-GAT framework and its components, and outline the evaluation metrics, baseline models, implementation details, and reproducibility considerations. Next, we report the experimental results and key insights, followed by a critical discussion of the findings, with emphasis on the performance of R-GAT relative to other models. Finally, the manuscript concludes with a summary of contributions and directions for future research.

\section{Literature Review}
\label{sec:literature}

Research on cancer text classification has progressed from traditional machine learning to deep learning and transformer-based approaches, with growing interest in graph-based methods. Early work primarily focused on single-domain resources such as radiology reports and clinical notes. For instance, Nguyen et al. \cite{nguyen2020hybrid} developed a hybrid encoder--decoder model with attention for Dutch radiology reports, achieving strong accuracy but facing challenges in clinical applicability. Similarly, Tang et al. \cite{Tang2019} fine-tuned BERT with an attention layer for clinical progress notes, attaining 97.6\% accuracy and underscoring the potential of transformers for medical text classification. More recent studies such as Uskaner et al. \cite{uskaner2023using} applied pre-trained BERT and DistilBERT to Turkish mammography reports, showing domain adaptation can yield high performance, with BERT achieving a 91\% F1-score.

Beyond cancer-specific tasks, graph-based neural architectures have gained traction in NLP and biomedical informatics. Ai et al. \cite{ai2024edge} proposed the Edge-Enhanced Minimum-Margin Graph Attention Network (EMGAN) to address short-text sparsity, while Wei et al. \cite{wei2024remaining} introduced a graph convolutional attention network (GCAN) with residual connections for time-series predictions. Other innovations, such as Song et al.’s Graph Sequence Pretraining with Transformer (GSPT) \cite{song2024pure} and Rao et al.’s Multi-layer Residual Attention Network (MRAN) \cite{rao2024knowledge}, highlight how attention and residual mechanisms can improve relational reasoning and semantic representation across diverse text-attributed graphs and knowledge graphs.

Together, these studies demonstrate the potential of attention-based and graph-based architectures to capture semantic and structural dependencies beyond linear text representations. However, most prior work has been limited to narrow contexts such as single cancer types, proprietary clinical notes, or imaging reports, leaving multi-cancer abstract classification comparatively underexplored. Moreover, while graph neural networks have been applied in biomedical and related domains, the role of residual connections within graph attention mechanisms has not been systematically examined for document-level biomedical classification, particularly under limited-data conditions.

This gap motivates the present work, which positions residual graph attention as a candidate solution for improving robustness and efficiency in cancer abstract classification.

\section{Dataset Construction and Statistics}
\label{sec:dataset-stat}
\subsection{Data Collection}

To provide a reliable benchmark for cancer document classification under limited-data conditions, a domain-specific dataset was constructed comprising 1,875 medical abstracts focusing on thyroid, colon, and lung cancers, as well as general biomedical topics. The abstracts were retrieved from PubMed, one of the largest biomedical literature databases, using the open-source Entrezpy \cite{buchmann2019entrezpy} Python library, which provides direct programmatic access to the NCBI Entrez system. Data collection was carried out between January and March 2024, and the search strategy was designed to incorporate cancer-specific terms such as ``thyroid cancer,'' ``colon cancer,'' and ``lung cancer,'' alongside more general biomedical keywords. Retrieval was restricted to articles published within the last five years to ensure that recent research trends were captured.  

An initial set of approximately 2,000 abstracts was obtained, from which non-English, duplicate, irrelevant, or incomplete entries were excluded. The remaining abstracts were manually reviewed and categorized into four groups: thyroid, colon, lung, and generic. The final dataset consisted of 1,875 unique abstracts with an average length of $\sim$145 tokens, ensuring both domain specificity and balanced topical coverage.

\subsection{Data Cleaning}
Prior to model development, the dataset underwent a systematic data cleaning and preprocessing pipeline to enhance quality and usability. Missing attributes were identified and resolved, followed by tokenization using the NLTK library \cite{bird2009nlp} to break text into smaller, meaningful units. To normalize word forms, lemmatization \cite{khyani2021interpretation} was applied, enabling the model to better capture semantic relationships. In addition, redundant and non-informative words were filtered out to reduce noise. Finally, the cleaned text was transformed into numerical vectors using multiple representation methods, including Term Frequency–Inverse Document Frequency (TF-IDF) \cite{dai2024ai}, Word2Vec embeddings \cite{goldberg2014word2vec}, and BERT-based tokenization \cite{devlin2019bertpretrainingdeepbidirectional}, enabling flexibility for downstream machine learning and deep learning tasks. This rigorous collection and cleaning procedure ensures that the dataset is both representative of recent biomedical literature and suitable for robust NLP-based experimentation.

\subsection{Dataset Statistics}
Table~\ref{tab:dataset_stats} summarizes the distribution of abstracts across the four categories together with the average number of tokens per abstract. The dataset is relatively well-balanced, with each category contributing between 450 and 480 abstracts. This balance reduces the likelihood of strong class imbalance effects, which can bias model training and evaluation. The average abstract length is approximately 145 tokens, though some variation exists across categories. For example, generic biomedical abstracts tend to be longer (163.72 tokens on average), while lung cancer abstracts are somewhat shorter (135.73 tokens on average). Such variation reflects the differing styles and focus of the source literature, but overall the abstracts remain concise and structurally comparable across classes.  

Although the dataset size (1,875 abstracts) may appear modest compared to large-scale general-domain corpora, it is consistent with established biomedical NLP resources, such as the BC5CDR corpus ($\sim$1,500 abstracts) \cite{li2016biocreative}. Domain-specific corpora in biomedical text mining are often smaller in scale because of the stringent filtering and manual curation required to ensure relevance, quality, and interpretability. The relatively balanced distribution of abstracts, coupled with careful preprocessing and annotation, ensures that this dataset is both representative of its target domain and suitable for benchmarking models under limited-data conditions.  

\begin{table}[ht]
\centering
    \caption{Distribution of abstracts across four cancer categories, including the number of documents and average tokens per abstract. The dataset is relatively balanced, with each class contributing a similar number of samples, helping to minimize class imbalance during evaluation.}

\label{tab:dataset_stats}
\begin{tabular}{lcc}
\toprule
\textbf{Category} & \textbf{No. of Abstracts} & \textbf{Avg. Tokens} \\
\midrule
Colon Cancer   & 468 & 144.45 \\
Generic        & 453 & 163.72 \\
Lung Cancer    & 473 & 135.73 \\
Thyroid Cancer & 481 & 136.00 \\
\midrule
\textbf{Total} & 1,875 & 144.74 \\
\bottomrule
\end{tabular}
\end{table}

\section{Methodology}
\label{sec:methodology}

\subsection{Overview of the Approach}

\begin{figure*}
\centering
\centerline{\includegraphics[width=\textwidth, height=8cm]{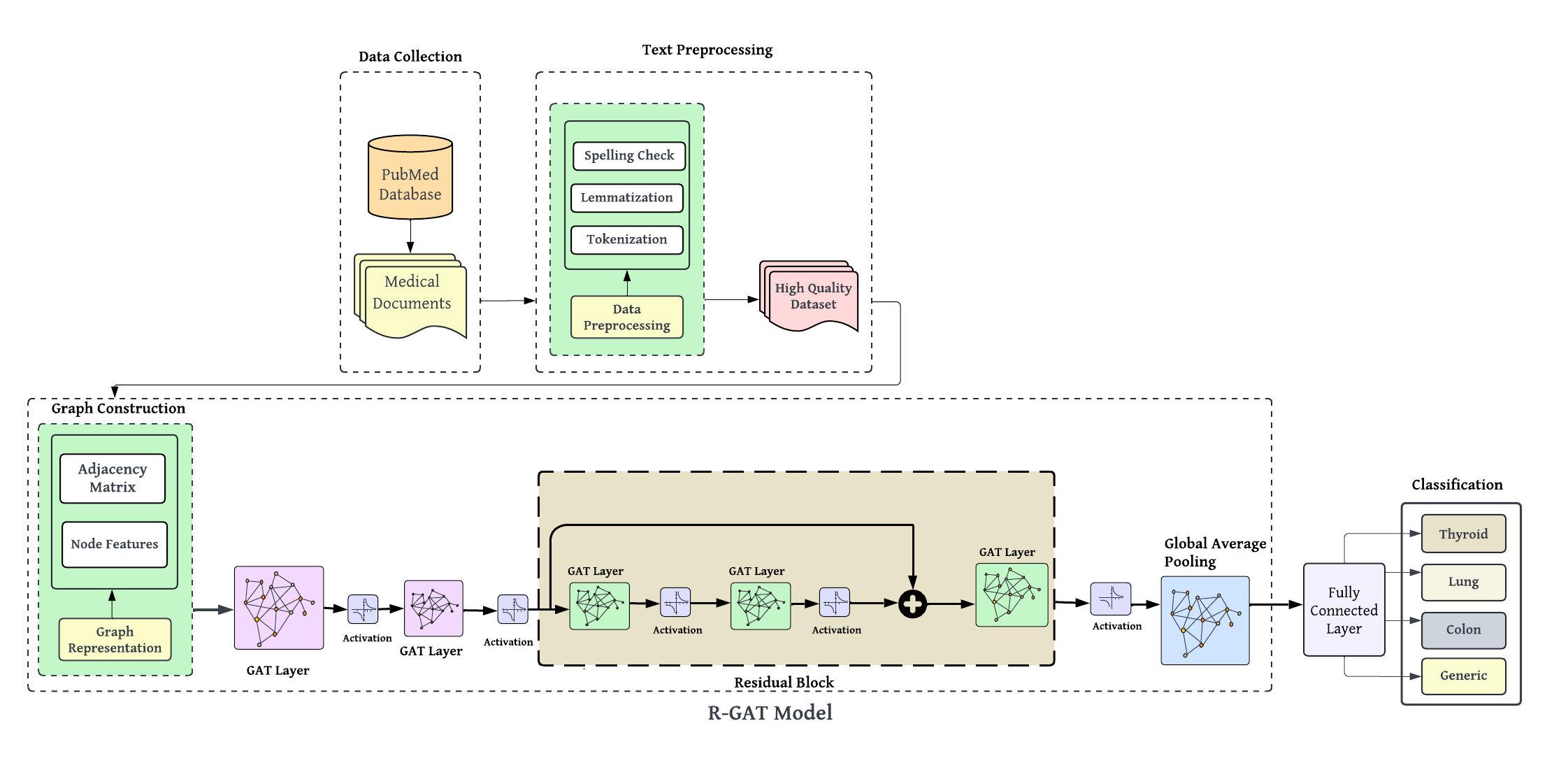}}

\caption{End-to-end methodology for cancer abstract classification using the proposed R-GAT. The workflow is divided into four major phases: (1) Data Collection: Abstracts are retrieved from PubMed and curated into a medical document corpus.
(2) Text Preprocessing: Cleaning operations include spelling correction, tokenization, and lemmatization, producing a high-quality dataset suitable for model training.
(3) Graph Construction and R-GAT Model Architecture: Abstracts are represented as graphs, where nodes correspond to document features and edges capture relational dependencies. The adjacency matrix and feature vectors form the graph representation. This representation is processed through stacked Graph Attention (GAT) layers with non-linear activations, followed by a Residual Block consisting of three GAT layers and skip connections. The residual design mitigates information loss and stabilizes training.
(4) Classification: Features are aggregated via a Global Average Pooling layer and passed through a fully connected layer with a Softmax decoder to predict four target categories: thyroid cancer, colon cancer, lung cancer, and generic biomedical abstracts.}

\label{fig:graphical-abstract}
\end{figure*}

This section outlines the classification of medical documents related to thyroid cancer, colon cancer, lung cancer, and generic topics divided into four subsequent phases. In the first phase, medical abstracts related to these cancers were collected from the PubMed database. The second phase involved text preprocessing, where the raw data underwent several techniques to produce a high-quality dataset, including tokenization, spelling checks, and text normalization, such as lemmatization. \\
The third phase included the R-GAT model, which unfolds across four distinct steps. Initially, a graph was constructed to represent node features and the adjacency matrix in the first step. In the second step, this graph was processed through two Graph Attention Network (GAT) layers before entering to the Residual Block. The third step introduced a Residual Block, a crucial component comprising three GAT layers, each with its activation function, as depicted in Fig. \ref{fig:graphical-abstract}. After that, in the forth step, a Global Average Pooling layer aggregated the features. Lastly, in the final phase of the workflow, a fully connected layer was used for classification.

\subsubsection{Graph Construction}

The first step is to create a graph that visually represents the cancer documents and their interconnections. In the node feature representation, each medical text document is assigned a feature vector to reflect its content. The feature matrix, denoted as $X \epsilon R$\textsuperscript{$N \times F$}, represents the document features, with $N$ representing the number of nodes in the graph and $F$ is the number of features per node.  An adjacency matrix $A \epsilon R$\textsuperscript{$N \times N$} is utilized to represent the relationships between documents, with $A$\textsubscript{$ij$} denoting the strength of the connection between document $i$ and document $j$. The edge weights can be acquired through training or determined using domain-specific knowledge.

\subsubsection{Graph Layers with Attention Mechanism}

A key component of our research is the use of the Graph Attention Mechanism, which computes attention scores for surrounding documents. The attention scores for a certain node $i$ are calculated as follows: We use Equation \ref{e1} to apply a Leaky Rectified Linear Unit (LeakyReLU) activation function to the concatenation of linear transformations of the feature vectors of nodes $i$ and $j$, where nodes $i$ and $j$ feature representations are denoted by $h_i$ and $h_j$, respectively; $a$ is a learnable attention weight vector and $W$ is a learnable weight matrix. The $(\top)$ represents that vector $a$ is transposed before performing the dot product to ensure appropriate dimension alignment for matrix multiplication. In addition, a double vertical bar sign denotes concatenation.

\begin{equation}
\hspace*{0.15\linewidth}
e_{ij} = \text{LeakyReLU}\left(a^{\top} \cdot \left[W \cdot h_i \| W \cdot h_j\right]\right)
\label{e1}
\end{equation}

To obtain attention coefficients, we normalize the attention scores using the SoftMax function in Equation \ref{e2}. In this regard, $\mathcal{N}(i)$ represents the collection of surrounding nodes of node $i$.

\begin{equation}
\hspace*{0.15\linewidth}
\alpha_{ij} = \frac{\exp \left(e_{ij}\right)}{\sum_{k \in \mathcal{N}(i)} \exp \left(e_{ik}\right)}
\label{e2}
\end{equation}

\subsubsection{Residual Blocks with Graph Attention Layers}

To improve our model and capture complex interactions, we use a Residual Block that combines multiple GAT layers, each followed by an activation function. The input $h_i$ in Equation \ref{e3} is the result of two previous GAT layers before the Residual Block. Equations \ref{e3}--\ref{e6} mathematically describe the structure of the Block, where $A$ represents the adjacency matrix. To be more precise, $h'_{i,1}$ and $h'_{i,2}$ represent the outputs of the first and second GAT layers, respectively; $h'_{i,3}$ is the result of adding the residual connection shown in (Equation \ref{e5}). Finally, $h'_{i}$ is the outcome of processing through the third GAT layer of the Residual Block.

\begin{equation}
\hspace*{0.15\linewidth}
h_{i,1}^{\prime} = \text{GATLayer}^1(h_{i},A)
\label{e3}
\end{equation}

\begin{equation}
\hspace*{0.15\linewidth}
h_{i,2}^{\prime} = \text{GATLayer}^2(h_{i,1}^{\prime},A)
\label{e4}
\end{equation}

\begin{equation}
\hspace*{0.15\linewidth}
h_{i,3}^{\prime} = h_{i} + h_{i,2}^{\prime} \quad \text{(Residual Connection)}
\label{e5}
\end{equation}

\begin{equation}
\hspace*{0.15\linewidth}
h_i^{\prime} = \text{GATLayer}^3(h_{i,3}^{\prime}, A)
\label{e6}
\end{equation}

The use of attention coefficients $\alpha_{ij}$ helps consolidate information from adjacent nodes, resulting in an improved representation for each particular node \(i\).

\begin{equation}
\hspace*{0.15\linewidth}
h_i^{\prime} = \sum_{j \in \mathcal{N}(i)} \alpha_{ij} \cdot W \cdot h_j
\label{e7}
\end{equation}

Simultaneously, to capture a wide range of patterns contained in the data, we employ several $K$ independent attention heads. Each attention head, $K$, which operates independently, captures different aspects of the interactions between nodes in the network. These different attention heads increase the model's ability to focus on diverse patterns at the same time. Also, the non-linear activation function, denoted as $\sigma$ further contributes to this process by introducing non-linearity, allowing the model to learn intricate relationships within data.

\begin{equation}
\hspace*{0.15\linewidth}
\vec{h}_{i}^{\prime} = \mathop{\|}_{k=1}^{K} \sigma\left(\sum_{j \in \mathcal{N}_{i}} \alpha_{i j}^{k} \mathbf{W}^{k} \vec{h}_{j}\right)
\label{e8}
\end{equation}

\subsubsection{Global Average Pooling Layer}

The final node representation is created by concatenating or averaging the results. Our network uses global average pooling, which computes the mean feature vector to represent the entire graph. Finally, the average feature vector passes through a dropout layer and then a fully connected layer, followed by a SoftMax activation function to forecast the cancer document classes. Algorithm \ref{alg:rgat_model} is the pseudocode of the R-GAT model, which illustrates the major steps and processes in the design.
\begin{algorithm}[htbp]
\caption{R-GAT Model}
\label{alg:rgat_model}
\footnotesize
\begin{algorithmic}[1]
\Require inputs
\State initialize \texttt{node\_feature\_matrix}, \texttt{adjacency\_matrix}
\State for each node in \texttt{node\_feature\_matrix} do
\State \quad calculate \texttt{attention\_scores} using Equation 1
\State \quad normalize \texttt{attention\_scores} using Equation 2
\State end for
\State for each Residual Block do
\State \quad for each GAT Layer do
\State \quad \quad update node feature representations using Equations 3--6
\State \quad end for
\State end for
\State aggregate node features using global average pooling and pass through fully connected layers
\State apply Softmax activation function for classification
\State return \texttt{classification\_output}
\end{algorithmic}
\end{algorithm}

\subsection{Decoder and Optimization Objective}

The final graph-level representation, obtained through global average pooling, is passed through a dropout layer and a fully connected decoder for classification. The decoder projects pooled embeddings into a vector with dimensionality equal to the number of cancer classes, followed by a softmax activation to produce class probabilities.

Training is optimized using categorical cross-entropy loss:
\[
\mathcal{L} = - \sum_{i=1}^{C} y_i \log(\hat{y}_i),
\]
where $C$ is the number of classes, $y_i$ is the ground-truth label, and $\hat{y}_i$ is the predicted probability. Optimization is performed with the Adam optimizer (learning rate = 0.001, weight decay = 0.0001), using early stopping based on validation macro-F1 to prevent overfitting.

\section{Experimental Setup}

\label{sec:exp-setup}

We benchmarked R-GAT against traditional machine learning, deep learning, and transformer-based baselines under stratified cross-validation protocols. Model performance was assessed using macro- and micro-averaged precision, recall, and F1-scores, with results reported as mean $\pm$ standard deviation to capture variability across folds. Confusion matrices were also generated to provide class-level error analysis, and 95\% confidence intervals were included where appropriate.

Full details of baseline architectures, feature extraction methods, training procedures, hyperparameter configurations, and the reproducibility statement are provided in the Appendix.

\section{Results and Analysis}
\label{sec:result-analysis}

\vspace{5pt}

\subsection{Insight 1: Performance of Baselines}
Traditional machine learning and deep learning models establish important reference points for evaluating R-GAT. As shown in Table~\ref{tab:ml_macro_micro}, Logistic Regression with TF-IDF (unigram) achieved a macro-F1 of $0.98 \pm 0.01$, underscoring the effectiveness of sparse lexical representations in small biomedical datasets. Gradient Boosting and AdaBoost also performed competitively, though with higher variance across folds. In contrast, Word2Vec embeddings consistently reduced performance (macro-F1 as low as 0.60), suggesting that dense embeddings without contextual information or domain adaptation are less effective in this setting.

Deep learning models (Table~\ref{tab:combined_deep_bert_rgat}) exhibited more variability. CNNs reached a macro-F1 of $0.96 \pm 0.01$, confirming their strength in capturing local semantic features. However, sequential models such as RNNs and shallow LSTMs underperformed substantially (macro-F1: 0.33–0.74), likely due to overfitting under limited data conditions. While transformer-based models (e.g., BioBERT, BioClinicalBERT) delivered the highest absolute scores ($0.98 \pm 0.00$), they require extensive computational resources and pretraining, making them less accessible for resource-constrained environments.

These baselines highlight two important themes: (1) lightweight linear models can yield surprisingly strong results, but their performance is highly dependent on feature design and may not generalize beyond TF-IDF representations; and (2) transformers achieve state-of-the-art accuracy, but at the cost of large compute requirements. These observations motivate the exploration of architectures such as R-GAT, which aim to balance accuracy, robustness, and efficiency under limited-data and limited-resource conditions.

\subsection{Insight 2: Robustness of R-GAT}

\begin{figure}[htb]
    \centering
    \begin{subfigure}[b]{0.48\textwidth}
        \centering
        \includegraphics[height=5cm]{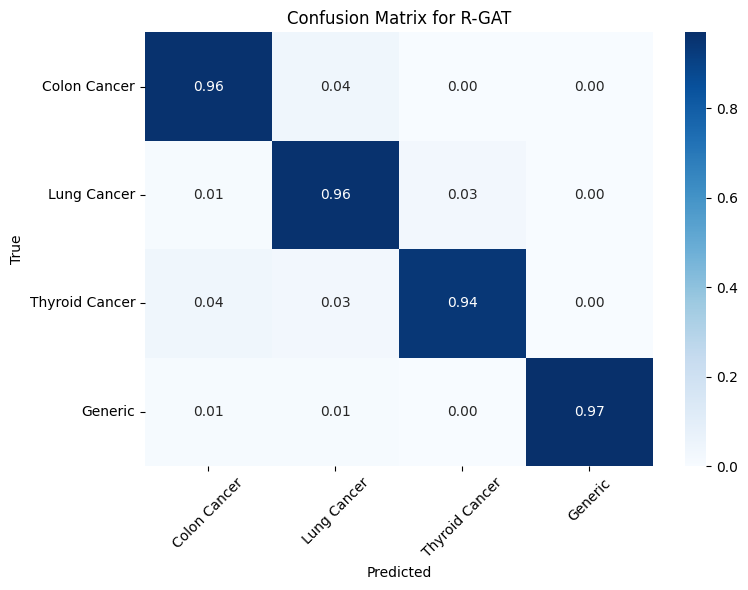}
        \caption{Confusion matrix of R-GAT across cancer classes.}
        \label{fig:confusion-matrix}
    \end{subfigure}
    \hfill
    \begin{subfigure}[b]{0.48\textwidth}
        \centering
        \includegraphics[height=5cm]{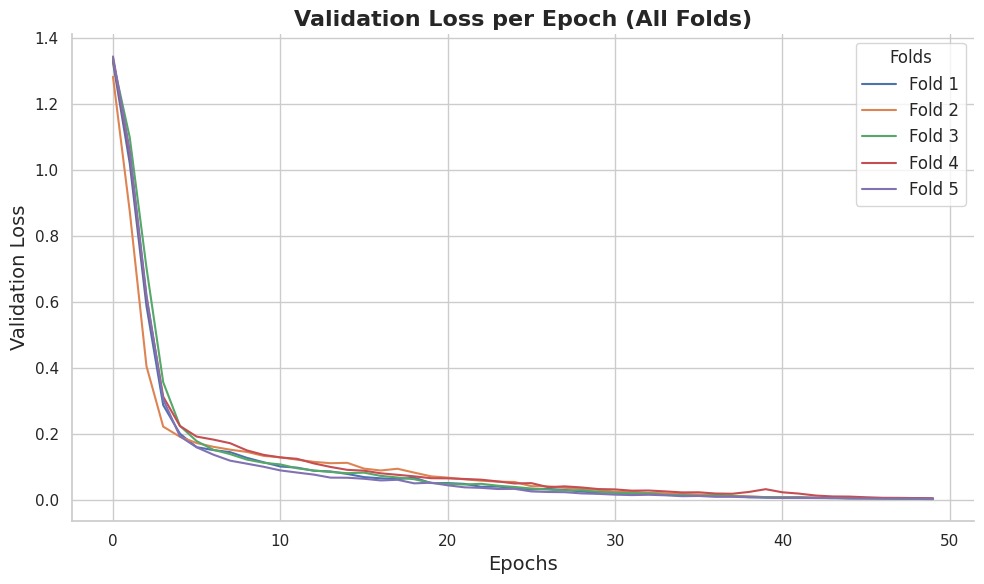}
        \caption{Training and validation loss across folds.}
        \label{fig:validation-loss}
    \end{subfigure}
    \caption{Performance visualization of the proposed R-GAT for multi-cancer abstract classification. (a) Confusion matrix showing the distribution of predictions across the four cancer classes: Colon Cancer, Lung Cancer, Thyroid Cancer, and Generic. Values on the diagonal represent correct classifications, with R-GAT achieving high accuracy across all categories ($\geq$0.94), indicating balanced performance and minimal class-specific bias. Off-diagonal values reflect misclassifications, which remain relatively rare. (b) Training and validation loss curves plotted over 50 epochs for each fold of stratified 5-fold cross-validation. The consistently smooth convergence across all folds demonstrates stable learning behavior and low variance, reinforcing the robustness of the R-GAT model under limited-data conditions.}
    \label{fig:combined-performance}
\end{figure}

\begin{figure}[htb]
    \centering
    \includegraphics[width=0.55\textwidth]{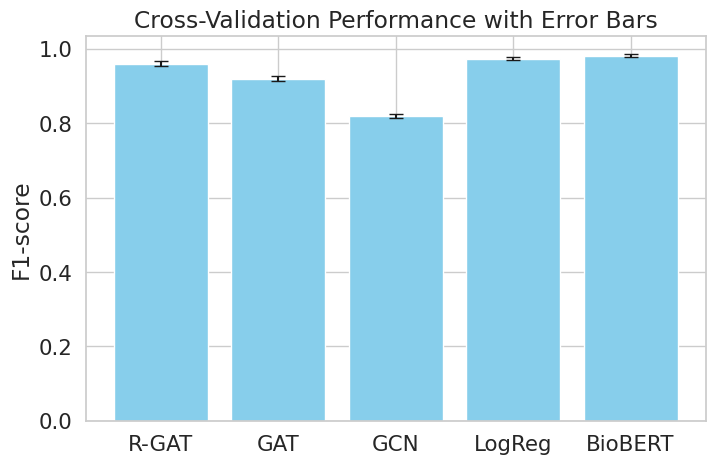}
    \caption{Cross-validation robustness analysis using F1-scores with 95\% confidence intervals (error bars) for R-GAT, its ablated variants (GAT without residuals and GCN without attention and residuals), and baseline models (Logistic Regression and BioBERT). R-GAT achieves a macro-F1 of approximately 0.96 with the narrowest confidence intervals, indicating strong stability and consistent generalization across folds. In contrast, GCN shows wider intervals and lower mean performance, reflecting higher sensitivity to data partitioning. Logistic Regression and BioBERT achieve slightly higher absolute scores, but with greater computational demands (BioBERT) or dependence on specific feature representations (LogReg). These results emphasize that R-GAT balances robustness and efficiency, making it particularly suitable for limited-data biomedical classification scenarios.}

    \label{fig:error-bar-performance}
\end{figure}

R-GAT delivered consistently strong results across cancer classes, achieving a macro-F1 of $0.96 \pm 0.01$. Figures~\ref{fig:combined-performance}(a) and \ref{fig:combined-performance}(b) illustrate stable convergence and balanced predictions across categories, while error bars in Figure~\ref{fig:error-bar-performance} confirm low variance under cross-validation. This contrasts with transformers and ensemble baselines, which showed greater fluctuations across folds.

To further assess robustness, we employed stratified 5-fold cross-validation with three independent random seeds, reducing the influence of any single data split. Table~\ref{tab:cv_results} compares three representative models: Logistic Regression with TF-IDF features, BioBERT, and the proposed R-GAT. Scores are reported as mean $\pm$ standard deviation across folds.  

\begin{table}[ht]
\centering
\caption{Cross-validation performance (mean $\pm$ std F1-score) for three representative models: Logistic Regression with TF-IDF features, BioBERT, and the proposed R-GAT, across four cancer classes. Results highlight both per-class and macro-averaged scores, allowing comparison of classical, transformer-based, and graph-based approaches under consistent evaluation.}

\label{tab:cv_results}
\begin{tabular}{lccccc}
\toprule
\textbf{Model} & \textbf{Colon} & \textbf{Generic} & \textbf{Lung} & \textbf{Thyroid} & \textbf{Macro F1} \\
\midrule
Logistic Regression (TF-IDF) & 0.97 $\pm$ 0.01 & 0.97 $\pm$ 0.01 & 0.98 $\pm$ 0.01 & 0.99 $\pm$ 0.00 & 0.98 $\pm$ 0.01 \\
BioBERT & 0.98 $\pm$ 0.00 & 0.98 $\pm$ 0.01 & 0.98 $\pm$ 0.01 & 0.99 $\pm$ 0.00 & 0.98 $\pm$ 0.00 \\
R-GAT (Proposed) & 0.95 $\pm$ 0.02 & 0.95 $\pm$ 0.02 & 0.97 $\pm$ 0.01 & 0.98 $\pm$ 0.01 & 0.96 $\pm$ 0.01 \\
\bottomrule
\end{tabular}
\end{table}

The results show a clear trade-off: Logistic Regression can slightly exceed R-GAT in absolute macro-F1, but its performance is highly dependent on TF-IDF features and may not extend to more complex representations. BioBERT achieves the strongest absolute scores but requires substantially greater computational resources and pretraining. In contrast, R-GAT maintains competitive accuracy while offering consistently low variance and efficiency in training. This positions R-GAT as a practical middle ground—balancing accuracy, robustness, and resource demands in biomedical NLP settings where stability is often more critical than marginal gains in peak performance.



\subsection{Insight 3: Contribution of Model Components}

Ablation experiments (Table \ref{tab:ablation_macro_micro}) underscore the critical role of residual connections and multi-head attention in the R-GAT architecture. Excluding residuals reduced macro-F1 to 0.92, while removing both residuals and attention caused a sharper decline to 0.83. These results validate the architectural choices, confirming that both mechanisms contribute substantially to robustness and generalization.

Overall, the evidence shows that although high-capacity models such as BioBERT achieve the strongest absolute scores, R-GAT delivers a more balanced trade-off—maintaining competitive accuracy while reducing variance and computational cost. This positions R-GAT as a lightweight yet reliable alternative for biomedical NLP tasks, particularly in scenarios constrained by data availability or computational resources.

\begin{table}[ht]
\centering
\caption{Macro- and micro-averaged precision, recall, and F1-scores for the full R-GAT model and its ablated variants. Results show the effect of removing residual connections (GAT) and both residuals and attention (GCN), providing a direct assessment of the contribution of each architectural component.}

\label{tab:ablation_macro_micro}
\begin{tabular}{lccc}
\toprule
\textbf{Model} & \textbf{Macro P / R / F1} & \textbf{Micro P / R / F1} \\
\midrule
R-GAT (full) & 0.97 / 0.96 / 0.96 & 0.97 / 0.96 / 0.96 \\
GAT (no residuals) & 0.92 / 0.93 / 0.92 & 0.92 / 0.93 / 0.92 \\
GCN (no attention, no residuals) & 0.83 / 0.83 / 0.82 & 0.83 / 0.83 / 0.82 \\
\bottomrule
\end{tabular}
\end{table}

\subsubsection{Inference Testing}

\label{inference-testing}

The predictive capability of R-GAT was further examined through inference testing on unseen biomedical abstracts (Fig. \ref{fig:inference-model}). In Fig. \ref{fig:inference-model} (a), the model correctly classified an abstract on the telomere–telomerase complex in familial and sporadic thyroid carcinoma as Thyroid Cancer. In Fig. \ref{fig:inference-model} (b), an abstract describing nitrosourea-based therapies for bronchogenic carcinoma was accurately labeled as Lung Cancer. These examples illustrate R-GAT’s ability to capture domain-specific terminology and contextual relationships, enabling precise predictions across distinct cancer types.

While inference outcomes alone cannot establish robustness, they demonstrate how the model translates learned representations into accurate real-world classifications. By leveraging residual graph attention, R-GAT effectively highlights key biomedical entities and preserves contextual dependencies, supporting generalization beyond the training set. This ability to maintain accuracy on new inputs underscores the model’s practical utility for biomedical text mining and its potential to enhance automated cancer literature classification.

\newtcolorbox{promptbox}{
  enhanced,
  colback=teal!10!white,
  colframe=gray!40!black,
  boxrule=0.7pt,
  arc=3pt,
  left=5pt,
  right=5pt,
  top=5pt,
  bottom=5pt,
  fonttitle=\bfseries,
  fontupper=\small,
  sharp corners=all,
  title style={left color=gray!15!white, right color=gray!5!white},
  coltitle=black,
}

\begin{figure*}[ht]
  \centering

  \begin{subfigure}[b]{0.9\textwidth}
    \centering
    \begin{promptbox}
      \textbf{Raw Abstract:} \textcolor{red}{Telomeres} are specialized structures at the ends of chromosomes, consisting of hundreds of repeated hexanucleotides (TTAGGG)n. Genetic integrity is partly maintained by the architecture of \textcolor{magenta}{telomeres}, and it is gradually lost as telomeres progressively shorten with each cell replication, due to incomplete lagging DNA strand synthesis and oxidative damage. \textcolor{red}{Telomerase} is a reverse transcriptase enzyme that counteracts telomere shortening by adding telomeric repeats to the G-rich strand. In the absence of telomerase or when the activity of the enzyme is low compared to the replicative erosion, \textcolor{red}{apoptosis} is triggered. Patients who have inherited genetic defects in telomere maintenance seem to have an increased risk of developing \textcolor{red}{malignant diseases}. At the somatic level, telomerase is reactivated in the majority of human carcinomas, suggesting that telomerase reactivation is a critical step for cancerogenesis. In sporadic thyroid carcinoma, telomerase activity is detectable in nearly 50\% of thyroid cancer tissues. Recently a germline alteration of the telomere-telomerase complex has been identified in patients with familial papillary thyroid cancer, characterized by short telomeres and increased expression and activity of telomerase compared to patients with sporadic papillary thyroid cancer. In this report, we review the role of the telomere-telomerase complex in sporadic and familial \textcolor{red}{thyroid cancer}.
      
      \vspace{0.3em}
      \textbf{Model Output:} \textcolor{red}{Thyroid Cancer}
    \end{promptbox}
    \caption{Abstract related to the role of the telomere-telomerase complex in familial and sporadic thyroid cancer.}
    \label{fig:thyroid-output}
  \end{subfigure}

  \vspace{1em}

  \begin{subfigure}[b]{0.9\textwidth}
    \centering
    \begin{promptbox}
      \textbf{Raw Abstract:} \textcolor{red}{BCNU}, \textcolor{red}{CCNU}, and \textcolor{red}{methyl-CCNU} have undergone extensive trials in multiple drug combinations for \textcolor{magenta}{bronchogenic carcinoma}. The addition of a \textcolor{red}{nitrosourea} appears to be an improvement over \textcolor{red}{cyclophosphamide} used alone in \textcolor{magenta}{oat cell carcinoma} and over the two-drug combination of cyclophosphamide and \textcolor{red}{methotrexate} in both \textcolor{magenta}{adenocarcinoma of the lung} and \textcolor{magenta}{oat cell disease}. Encouraging response rates have been seen in \textcolor{magenta}{squamous lung cancer} with multiple-drug combinations of a nitrosourea, an alkylating agent, \textcolor{red}{vincristine}, and \textcolor{red}{bleomycin} with or without \textcolor{red}{adriamycin}. The nitrosoureas have been easily incorporated, at reduced doses, into multiple-drug regimens with cumulative \textcolor{red}{myelosuppression} seen only when the interval between nitrosourea doses is less than 6 weeks. Conclusions about the ultimate role of these compounds in \textcolor{magenta}{lung cancer} treatment must await (a) comparative trials of combinations with and without a nitrosourea, and (b) further exploration of new approaches to increase their therapeutic index.
      
      \vspace{0.3em}
      \textbf{Model Output:} \textcolor{red}{Lung Cancer}
    \end{promptbox}
    \caption{Abstract discussing the effectiveness of nitrosoureas in treating lung cancer.}
    \label{fig:lung-output}
  \end{subfigure}

  \caption{Analysis of cancer abstracts fed into the R-GAT model for classification: 
  (a) Thyroid Cancer—the model analyzed the abstract focused on the telomere-telomerase complex in both sporadic and familial thyroid cancer cases, emphasizing telomere shortening and telomerase activation; 
  (b) Lung Cancer—the model processed an abstract detailing the effectiveness of nitrosoureas and other agents in treating various types of lung cancer, including oat cell carcinoma and adenocarcinoma. 
  Both abstracts were correctly classified by the R-GAT model.}
  \label{fig:inference-model}
\end{figure*}

\subsubsection{Comparative Review of Existing Studies}
\label{compare-research}

Table~\ref{table:cancer-classification} presents a comparative summary of studies that have employed various data sources and techniques for cancer classification across types such as breast, colorectal, prostate, lung carcinoma, thyroid, colon, and lung. As shown in the table, most of the datasets used in prior studies were not publicly available, indicating limited accessibility and reproducibility. Additionally, the “Multi-cancer” column highlights that the majority of studies focused on a single cancer type, with only a few addressing multiple types. This reveals a key limitation, as studying multiple cancers together can enable more generalized and comprehensive models. Previous research has largely concentrated on radiological and clinical reports, particularly for breast cancer. In contrast, biomedical abstracts have been considerably underexplored, despite their availability and potential to reveal diverse clinical insights. These abstracts offer a rich source of semantic information that can support both single and multi-cancer classification tasks.

Although some prior studies have incorporated transformer-based models, Table~\ref{table:cancer-classification} shows that graph-based attention networks have not yet been applied for cancer classification using biomedical abstracts. Moreover, none of the reviewed works specifically targeted thyroid, colon, and lung cancers within this modality. To address these gaps, the R-GAT was designed to classify lengthy medical abstracts by capturing semantic and relational patterns within text using graph-based representations. The performance of the R-GAT was evaluated against state-of-the-art transformer models, including BERT, BioBERT, RoBERTa, and Bio+ClinicalBERT, as well as traditional machine learning and deep learning methods. Experimental results demonstrate that R-GAT achieves strong generalizability across cancer types and outperforms existing models in classifying biomedical abstracts.

\section{Discussion}
\label{sec:discussion}

Across the benchmarking spectrum, a consistent pattern emerged: simple models such as Logistic Regression and resource-intensive transformers like BioBERT both achieved strong performance on this dataset, yet their behavior proved less stable when examined across folds and feature variations. This underscores a broader issue in biomedical NLP—absolute scores alone do not capture the practical reliability of a model, especially under low-data constraints. What matters is not only peak accuracy but also how consistently a model generalizes across different conditions.

R-GAT’s strength lies in this dimension of stability. By combining multi-head attention with residual connections, the model maintained balanced performance across cancer categories and resisted fluctuations introduced by different data partitions. While it did not surpass all baselines in raw F1, R-GAT demonstrated that graph-based architectures can reliably capture relational dependencies in text without relying on large-scale pretraining or extensive computational resources. In practice, this makes R-GAT a dependable option in biomedical environments where annotated data are scarce and hardware resources are limited.

The dataset itself imposes certain constraints. Its modest size and focus on single-topic abstracts simplify the classification task and reduce the diversity of linguistic patterns available for training. This explains why even lightweight baselines perform unexpectedly well and why high scores should not be over-interpreted as evidence of clinical applicability. Recognizing these limits avoids overstating contributions, while still emphasizing the value of providing a clean, balanced corpus for reproducible cancer informatics research.

The findings position R-GAT as a complementary approach within the biomedical NLP landscape. Transformers remain unmatched in accuracy when large-scale data and compute are available, whereas linear models offer competitive baselines for sparse feature spaces. Between these extremes, R-GAT provides a middle ground: a lightweight, interpretable architecture that delivers stable performance under realistic constraints. Its role is less about replacing existing methods and more about broadening the methodological toolkit for scenarios where robustness and efficiency are as critical as raw accuracy.

\section{Conclusion}
\label{sec:conclusion}

This study introduced R-GAT, a residual graph attention network developed for cancer abstract classification under limited-data conditions. Through systematic benchmarking against transformer-based and traditional baselines, R-GAT was shown to achieve competitive accuracy, reduced variance across folds, and efficiency advantages in computationally constrained environments. In addition, a curated dataset of 1,875 PubMed abstracts was released to facilitate reproducibility and provide a standardized benchmark for future investigations.

Nevertheless, the study has certain limitations. Specifically, the dataset is modest in size and has not been clinically validated, which restricts direct applicability in oncology practice. Looking ahead, future work should focus on extending the framework to additional cancer types, incorporating multi-modal data sources, and exploring hybrid graph–transformer architectures.

Taken in combination, the release of both the model and dataset provides a reproducible foundation for cancer informatics research and supplies a resource that can be integrated into benchmarking efforts and community challenges. Consequently, the findings position R-GAT as a robust and efficient alternative for biomedical NLP tasks in data-constrained settings.

\section*{Declaration of Competing Interest}

The authors declare that the research was conducted in the
absence of any commercial or financial relationships that could
be construed as a potential conflict of interest.

\section*{Funding Declaration}

This research did not receive any specific grant from funding agencies in the public, commercial, or not-for-profit sectors.

\section*{Author Contributions Statement}

E.H. conceived the research idea, designed the experiments, implemented the codebase, and wrote the full draft of the manuscript. T.N. contributed to refining the experimental results, enhancing performance evaluation, and improving the illustrations. S.M., S.R., and N.A.G. supervised the research work and provided critical feedback throughout the project. All authors reviewed and approved the final manuscript.


\bibliography{sample}
\bibliographystyle{unsrtnat}

\appendix
\section{Appendix}

\renewcommand{\thetable}{A\arabic{table}}
\setcounter{table}{0}

\tableofcontents

\subsection{Experimental Details}
\label{appendix:experimental-details}

To support reproducibility, we provide detailed descriptions of baseline models, feature extraction methods, training procedures, and hyperparameter settings that complement the main Experimental Setup.  

\subsubsection{Baseline Models}

To ensure a comprehensive and fair evaluation of the proposed R-GAT framework, we benchmarked against a wide spectrum of baseline models that reflect three major generations of NLP research. These include:  

\textbf{Traditional Machine Learning.}  
Logistic Regression, Random Forest, Support Vector Machine (SVM), Multinomial Naïve Bayes, K-Nearest Neighbors, Gradient Boosting, AdaBoost, Decision Trees, and XGBoost. These models remain popular in biomedical text classification due to their simplicity, interpretability, and strong performance with sparse features such as TF-IDF.  

\textbf{Deep Learning Architectures.}  
Convolutional Neural Networks (CNN), Recurrent Neural Networks (RNN), Long Short-Term Memory (LSTM), Gated Recurrent Units (GRU), Bidirectional LSTM, Stacked variants, and a Hybrid CNN–BiLSTM ensemble. These models are widely used to capture sequential or local dependencies in biomedical texts.  

\textbf{Transformer-based Models.}  
BERT, BioBERT, RoBERTa, and BioClinicalBERT. These pre-trained architectures provide contextual embeddings and represent the current state-of-the-art in biomedical NLP, making them essential baselines for comparison. Together, these baselines establish a rigorous evaluation spectrum: from interpretable statistical methods, to sequence-based neural networks, to large-scale pretrained transformers. Detailed feature extraction methods for these baselines are described in Section~A.1.2.

\subsubsection{Feature Extraction Methods}

To ensure a fair and consistent evaluation across all baselines, we employed a diverse set of feature extraction methods ranging from classical frequency-based representations to modern contextual tokenizers. This diversity allowed us to assess how different feature spaces influence the performance of traditional machine learning, deep learning, and transformer-based approaches in cancer document classification.

For traditional models, we used Term Frequency–Inverse Document Frequency (TF-IDF) with both unigram and bigram configurations. TF-IDF highlights terms that are distinctive within individual documents relative to the corpus, making it a strong choice for sparse, high-dimensional representations that work well with classifiers such as Logistic Regression and SVM. To provide dense semantic embeddings, we incorporated Word2Vec, which maps words into continuous vector spaces based on contextual similarity. These embeddings are particularly valuable for deep learning models like CNNs and LSTMs, as they capture distributional semantics beyond raw frequency counts.

For several deep learning baselines, including CNNs, RNNs, LSTMs, and GRUs, we also utilized Keras embedding layers. These embeddings, trained jointly with the model, offer task-specific representations and provide a strong baseline when external pretrained embeddings are not applied. Finally, for transformer-based baselines, we adopted the native tokenizers of BERT, BioBERT, RoBERTa, and Bio+ClinicalBERT. These tokenizers segment text into subword units according to their pretraining strategies, ensuring proper handling of biomedical terminology and domain-specific abbreviations. Aligning the input representation with each model’s pretraining method allowed for optimal evaluation of transformer-based baselines.

Taken together, these feature extraction methods represent three complementary categories: sparse frequency-based features (TF-IDF), dense distributional embeddings (Word2Vec and Keras embeddings), and contextualized subword tokenization (transformer tokenizers). This balanced set of approaches ensures that our comparisons are rigorous and reflect the full spectrum of feature representation strategies in biomedical NLP.

\subsubsection{Training Procedure}

All experiments were conducted under a consistent training and evaluation protocol to ensure reproducibility. For the machine learning baselines, we used stratified $k$-fold cross-validation to evaluate model performance. Specifically, a 5-fold stratified split repeated three times (15 runs in total) was employed for traditional ML models, while deep learning and transformer-based models were evaluated using a 3-fold stratified cross-validation due to higher computational cost. For each fold, models were trained on the training set and evaluated on the held-out test fold, with results reported as the mean ± standard deviation across folds.

To reduce variance due to random initialization, each experiment was repeated with multiple random seeds (three runs per setting), and averaged results are reported. Early stopping was applied to deep learning and R-GAT models with a patience of 10 epochs, monitored on the validation macro-F1 score. Transformer-based models were fine-tuned for a fixed number of epochs (two) following standard practice, without additional early stopping. This procedure ensures that the reported results reflect stable performance estimates rather than isolated runs, and that conclusions are robust across different data splits and initialization seeds.

\subsubsection{Hyperparameter Settings}

To ensure reproducibility and transparency, we summarize the hyperparameter configurations used for all models, including traditional machine learning baselines, deep learning architectures, transformer-based models, and the proposed R-GAT framework.

\textbf{Machine Learning Models.}  
We evaluated multiple classifiers including Decision Tree, Random Forest, Gradient Boosting, AdaBoost, K-Nearest Neighbors (KNN), Logistic Regression, Support Vector Machine (SVM with linear kernel), Naïve Bayes (MultinomialNB and GaussianNB), and XGBoost. For text representation, both TF-IDF (unigram and bigram) and Word2Vec embeddings (100-dimensional GloVe pre-trained vectors) were employed. Most classifiers were used with scikit-learn default hyperparameters, with Logistic Regression configured with a maximum of 500 iterations and XGBoost trained with \texttt{eval\_metric=mlogloss}. Repeated stratified 5-fold cross-validation with three repeats was used for evaluation.

\textbf{Deep Learning Models (non-transformer).}  
Neural architectures were implemented in Keras/TensorFlow with a vocabulary size of 20,000 and sequence length fixed to 200 tokens. All models used 128-dimensional embeddings. Architectures included CNN, RNN, LSTM, GRU, Bi-LSTM, Stacked LSTM, Stacked Bi-LSTM, and a Hybrid CNN–BiLSTM ensemble. Each network was trained with categorical cross-entropy loss and the Adam optimizer, using a batch size of 32 and up to 2 epochs per fold (for cross-validation). Dropout layers were included where appropriate to prevent overfitting.

\textbf{Transformer Models.} 
We fine-tuned four pretrained transformer architectures. The key configurations are summarized below:

\begin{itemize}
    \item \textbf{BERT} (\texttt{bert-base-uncased})
    \item \textbf{BioBERT} (\texttt{dmis-lab/biobert-base-cased-v1.1})
    \item \textbf{RoBERTa} (\texttt{roberta-base})
    \item \textbf{BioClinicalBERT} (\texttt{emilyalsentzer/Bio\_ClinicalBERT})
\end{itemize}

Tokenization was carried out using Hugging Face tokenizers with a maximum sequence length of 200. 
All models were fine-tuned for two epochs via the \texttt{Trainer} API using the AdamW optimizer (default hyperparameters), cross-entropy loss, a batch size of 8, and a fixed random seed of 42. 
Three-fold stratified cross-validation was employed for evaluation. 
To ensure reproducibility and prevent variation across folds, external logging and checkpoint saving were disabled.

\textbf{Proposed R-GAT Model.}  
The R-GAT was implemented with a hidden dimension of 128, four attention heads per layer, and five graph attention layers. A dropout rate of 0.3 was applied after each layer. The network employed global average pooling, followed by a fully connected softmax classification layer. Training used the Adam optimizer with a learning rate of 0.001 and weight decay of 0.0001, for 200 epochs with a batch size of 32. Early stopping with a patience of 10 epochs was applied based on validation macro-F1 score. All R-GAT experiments were conducted on an NVIDIA A100 GPU within the Google Colab environment.

\subsubsection{Reproducibility Details}

All experiments were conducted on an NVIDIA A100 GPU (40 GB memory) within the Google Colab Pro+ environment. Machine learning and deep learning baselines were implemented using Python 3.12, scikit-learn (v1.5), and TensorFlow/Keras (v2.16), while transformer-based models were fine-tuned using PyTorch (v2.2) and HuggingFace Transformers (v4.41). The proposed R-GAT model was implemented using the Deep Graph Library (DGL, v1.1) with PyTorch as backend. Approximate training time per model was 5–10 minutes for machine learning baselines, 20–40 minutes for deep learning architectures, 1–2 hours for transformer-based models, and 2–3 hours for R-GAT on the full dataset. To support reproducibility, a subset of the dataset and code has been publicly released and is accessible here:  
\url{https://github.com/eliashossain001/MedicalAbstracts/blob/main/MedicalAbstracts.csv}

\subsection{Extended Results for Baseline Models}
Table~\ref{tab:ml_macro_micro} reports macro and micro averages of precision, recall, and F1-score for traditional machine learning models under different feature extraction methods (TF-IDF unigrams/bigrams, Word2Vec). Results are reported as mean $\pm$ standard deviation across stratified 5-fold cross-validation.

\renewcommand{\thetable}{A\arabic{table}}
\setcounter{table}{0} 

\begin{table*}[ht]
\centering
\caption{Macro- and micro-averaged precision, recall, and F1-scores for conventional machine learning models under different feature extraction strategies: TF-IDF (unigram and bigram) and Word2Vec embeddings. Results are reported as mean $\pm$ standard deviation across stratified 5-fold cross-validation. The table highlights how sparse TF-IDF features generally yield stronger performance across models, while dense Word2Vec embeddings lead to reduced accuracy, reflecting the influence of feature representation on classical classifiers.}

\label{tab:ml_macro_micro}
\resizebox{\textwidth}{!}{%
\begin{tabular}{l|l|ccc|ccc}
\toprule
\multirow{2}{*}{\textbf{Model}} & \multirow{2}{*}{\textbf{Feature}} & \multicolumn{3}{c|}{\textbf{Macro}} & \multicolumn{3}{c}{\textbf{Micro}} \\
 & & \textbf{Precision} & \textbf{Recall} & \textbf{F1} & \textbf{Precision} & \textbf{Recall} & \textbf{F1} \\
\midrule
Decision Tree & TF-IDF (Unigram) & 0.96$\pm$0.01 & 0.96$\pm$0.01 & 0.96$\pm$0.01 & 0.96$\pm$0.01 & 0.96$\pm$0.01 & 0.96$\pm$0.01 \\
 & TF-IDF (Bigram) & 0.94$\pm$0.01 & 0.93$\pm$0.01 & 0.94$\pm$0.01 & 0.93$\pm$0.01 & 0.93$\pm$0.01 & 0.93$\pm$0.01 \\
 & Word2Vec & 0.60$\pm$0.02 & 0.59$\pm$0.02 & 0.60$\pm$0.02 & 0.59$\pm$0.02 & 0.59$\pm$0.02 & 0.59$\pm$0.02 \\
\midrule
Random Forest & TF-IDF (Unigram) & 0.98$\pm$0.01 & 0.98$\pm$0.01 & 0.98$\pm$0.01 & 0.98$\pm$0.01 & 0.98$\pm$0.01 & 0.98$\pm$0.01 \\
 & TF-IDF (Bigram) & 0.96$\pm$0.01 & 0.95$\pm$0.01 & 0.95$\pm$0.01 & 0.95$\pm$0.01 & 0.95$\pm$0.01 & 0.95$\pm$0.01 \\
 & Word2Vec & 0.80$\pm$0.02 & 0.80$\pm$0.02 & 0.80$\pm$0.02 & 0.80$\pm$0.02 & 0.80$\pm$0.02 & 0.80$\pm$0.02 \\
\midrule
KNN & TF-IDF (Unigram) & 0.90$\pm$0.02 & 0.90$\pm$0.02 & 0.90$\pm$0.02 & 0.90$\pm$0.02 & 0.90$\pm$0.02 & 0.90$\pm$0.02 \\
 & TF-IDF (Bigram) & 0.91$\pm$0.01 & 0.91$\pm$0.01 & 0.91$\pm$0.01 & 0.91$\pm$0.01 & 0.91$\pm$0.01 & 0.91$\pm$0.01 \\
 & Word2Vec & 0.74$\pm$0.02 & 0.74$\pm$0.02 & 0.74$\pm$0.02 & 0.74$\pm$0.02 & 0.74$\pm$0.02 & 0.74$\pm$0.02 \\
\midrule
Multinomial NB & TF-IDF (Unigram) & 0.94$\pm$0.01 & 0.94$\pm$0.01 & 0.94$\pm$0.01 & 0.94$\pm$0.01 & 0.94$\pm$0.01 & 0.94$\pm$0.01 \\
 & TF-IDF (Bigram) & 0.93$\pm$0.01 & 0.92$\pm$0.01 & 0.92$\pm$0.01 & 0.92$\pm$0.01 & 0.92$\pm$0.01 & 0.92$\pm$0.01 \\
 & Word2Vec & 0.74$\pm$0.02 & 0.74$\pm$0.02 & 0.74$\pm$0.02 & 0.73$\pm$0.02 & 0.73$\pm$0.02 & 0.73$\pm$0.02 \\
\midrule
Gradient Boosting & TF-IDF (Unigram) & 0.97$\pm$0.01 & 0.97$\pm$0.01 & 0.97$\pm$0.01 & 0.97$\pm$0.01 & 0.97$\pm$0.01 & 0.97$\pm$0.01 \\
 & TF-IDF (Bigram) & 0.95$\pm$0.01 & 0.94$\pm$0.01 & 0.94$\pm$0.01 & 0.94$\pm$0.01 & 0.94$\pm$0.01 & 0.94$\pm$0.01 \\
 & Word2Vec & 0.85$\pm$0.02 & 0.85$\pm$0.02 & 0.85$\pm$0.02 & 0.85$\pm$0.02 & 0.85$\pm$0.02 & 0.85$\pm$0.02 \\
\midrule
AdaBoost & TF-IDF (Unigram) & 0.96$\pm$0.01 & 0.96$\pm$0.01 & 0.96$\pm$0.01 & 0.96$\pm$0.01 & 0.96$\pm$0.01 & 0.96$\pm$0.01 \\
 & TF-IDF (Bigram) & 0.93$\pm$0.01 & 0.91$\pm$0.01 & 0.91$\pm$0.01 & 0.91$\pm$0.01 & 0.91$\pm$0.01 & 0.91$\pm$0.01 \\
 & Word2Vec & 0.69$\pm$0.02 & 0.68$\pm$0.02 & 0.69$\pm$0.02 & 0.68$\pm$0.02 & 0.68$\pm$0.02 & 0.68$\pm$0.02 \\
\midrule
SVM & TF-IDF (Unigram) & 0.98$\pm$0.01 & 0.98$\pm$0.01 & 0.98$\pm$0.01 & 0.98$\pm$0.01 & 0.98$\pm$0.01 & 0.98$\pm$0.01 \\
 & TF-IDF (Bigram) & 0.95$\pm$0.01 & 0.93$\pm$0.01 & 0.94$\pm$0.01 & 0.93$\pm$0.01 & 0.93$\pm$0.01 & 0.93$\pm$0.01 \\
 & Word2Vec & 0.89$\pm$0.01 & 0.89$\pm$0.01 & 0.89$\pm$0.01 & 0.89$\pm$0.01 & 0.89$\pm$0.01 & 0.89$\pm$0.01 \\
\midrule
XGBoost & TF-IDF (Unigram) & 0.97$\pm$0.01 & 0.97$\pm$0.01 & 0.97$\pm$0.01 & 0.97$\pm$0.01 & 0.97$\pm$0.01 & 0.97$\pm$0.01 \\
 & TF-IDF (Bigram) & 0.94$\pm$0.01 & 0.94$\pm$0.01 & 0.94$\pm$0.01 & 0.94$\pm$0.01 & 0.94$\pm$0.01 & 0.94$\pm$0.01 \\
 & Word2Vec & 0.85$\pm$0.02 & 0.85$\pm$0.02 & 0.85$\pm$0.02 & 0.85$\pm$0.02 & 0.85$\pm$0.02 & 0.85$\pm$0.02 \\
\midrule
Logistic Regression & TF-IDF (Unigram) & 0.98$\pm$0.01 & 0.98$\pm$0.01 & 0.98$\pm$0.01 & 0.98$\pm$0.01 & 0.98$\pm$0.01 & 0.98$\pm$0.01 \\
 & TF-IDF (Bigram) & 0.95$\pm$0.01 & 0.94$\pm$0.01 & 0.94$\pm$0.01 & 0.94$\pm$0.01 & 0.94$\pm$0.01 & 0.94$\pm$0.01 \\
 & Word2Vec & 0.89$\pm$0.01 & 0.89$\pm$0.01 & 0.89$\pm$0.01 & 0.89$\pm$0.01 & 0.89$\pm$0.01 & 0.89$\pm$0.01 \\
\bottomrule
\end{tabular}%
}
\end{table*}

\subsection{Extended Results for Deep Learning and Transformer Models}
Table~\ref{tab:combined_deep_bert_rgat} summarizes performance across deep learning models, BERT-family baselines, and the proposed R-GAT with its ablations. This extended comparison complements the main text by showing consistency across evaluation protocols.

\begin{table*}[ht]
\centering
\caption{Macro- and micro-averaged precision, recall, and F1-scores for deep learning models (CNN, RNN, LSTM variants, GRU), transformer-based baselines (BERT, BioBERT, RoBERTa, BioClinicalBERT), and the proposed R-GAT with ablated variants. Results are reported as mean $\pm$ standard deviation across stratified 5-fold cross-validation. The table illustrates that while transformer models achieve near-perfect performance on this dataset, they are computationally expensive, whereas R-GAT delivers competitive results with lower variance. The ablation rows highlight the individual contributions of residual connections and attention mechanisms to overall model robustness.}

\label{tab:combined_deep_bert_rgat}
\resizebox{\textwidth}{!}{%
\begin{tabular}{l|ccc|ccc}
\toprule
\multirow{2}{*}{\textbf{Model}} & \multicolumn{3}{c|}{\textbf{Macro}} & \multicolumn{3}{c}{\textbf{Micro}} \\
 & \textbf{Precision} & \textbf{Recall} & \textbf{F1} & \textbf{Precision} & \textbf{Recall} & \textbf{F1} \\
\midrule
CNN & 0.96 $\pm$ 0.01 & 0.96 $\pm$ 0.01 & 0.96 $\pm$ 0.01 & 0.96 $\pm$ 0.01 & 0.96 $\pm$ 0.01 & 0.96 $\pm$ 0.01 \\
RNN & 0.38 $\pm$ 0.03 & 0.36 $\pm$ 0.02 & 0.33 $\pm$ 0.03 & 0.36 $\pm$ 0.02 & 0.36 $\pm$ 0.02 & 0.36 $\pm$ 0.02 \\
LSTM & 0.78 $\pm$ 0.05 & 0.74 $\pm$ 0.06 & 0.71 $\pm$ 0.08 & 0.74 $\pm$ 0.06 & 0.74 $\pm$ 0.06 & 0.74 $\pm$ 0.06 \\
GRU & 0.74 $\pm$ 0.03 & 0.73 $\pm$ 0.03 & 0.73 $\pm$ 0.03 & 0.73 $\pm$ 0.03 & 0.73 $\pm$ 0.03 & 0.73 $\pm$ 0.03 \\
Bi-LSTM & 0.73 $\pm$ 0.07 & 0.65 $\pm$ 0.05 & 0.65 $\pm$ 0.06 & 0.65 $\pm$ 0.06 & 0.65 $\pm$ 0.06 & 0.65 $\pm$ 0.06 \\
Stacked LSTM & 0.81 $\pm$ 0.03 & 0.79 $\pm$ 0.04 & 0.78 $\pm$ 0.04 & 0.79 $\pm$ 0.04 & 0.79 $\pm$ 0.04 & 0.79 $\pm$ 0.04 \\
Stacked Bi-LSTM & 0.82 $\pm$ 0.05 & 0.77 $\pm$ 0.09 & 0.74 $\pm$ 0.13 & 0.77 $\pm$ 0.09 & 0.77 $\pm$ 0.09 & 0.77 $\pm$ 0.09 \\
Hybrid Ensemble & 0.81 $\pm$ 0.08 & 0.79 $\pm$ 0.09 & 0.78 $\pm$ 0.10 & 0.79 $\pm$ 0.09 & 0.79 $\pm$ 0.09 & 0.79 $\pm$ 0.09 \\
\midrule
BioBERT & 0.98 $\pm$ 0.00 & 0.98 $\pm$ 0.00 & 0.98 $\pm$ 0.00 & 0.98 $\pm$ 0.00 & 0.98 $\pm$ 0.00 & 0.98 $\pm$ 0.00 \\
BERT & 0.98 $\pm$ 0.00 & 0.98 $\pm$ 0.00 & 0.98 $\pm$ 0.00 & 0.98 $\pm$ 0.00 & 0.98 $\pm$ 0.00 & 0.98 $\pm$ 0.00 \\
RoBERTa & 0.97 $\pm$ 0.00 & 0.97 $\pm$ 0.00 & 0.97 $\pm$ 0.00 & 0.97 $\pm$ 0.00 & 0.97 $\pm$ 0.00 & 0.97 $\pm$ 0.00 \\
BioClinicalBERT & 0.98 $\pm$ 0.00 & 0.98 $\pm$ 0.00 & 0.98 $\pm$ 0.00 & 0.98 $\pm$ 0.00 & 0.98 $\pm$ 0.00 & 0.98 $\pm$ 0.00 \\
\midrule
R-GAT (full) & 0.97 $\pm$ 0.01 & 0.96 $\pm$ 0.01 & 0.96 $\pm$ 0.01 & 0.97 $\pm$ 0.01 & 0.96 $\pm$ 0.01 & 0.96 $\pm$ 0.01 \\
GAT (no residuals) & 0.92 $\pm$ 0.01 & 0.93 $\pm$ 0.01 & 0.92 $\pm$ 0.01 & 0.92 $\pm$ 0.01 & 0.93 $\pm$ 0.01 & 0.92 $\pm$ 0.01 \\
GCN (no attention, no residuals) & 0.83 $\pm$ 0.01 & 0.83 $\pm$ 0.01 & 0.82 $\pm$ 0.01 & 0.83 $\pm$ 0.01 & 0.83 $\pm$ 0.01 & 0.82 $\pm$ 0.01 \\
\bottomrule
\end{tabular}%
}
\end{table*}

\subsection{Comparative Analysis with Prior Work}
For completeness, Table~\ref{table:cancer-classification} compares prior research on cancer classification based on cancer type, dataset type, public availability, multi-cancer coverage, and transformer usage. This contextualizes the novelty of R-GAT within the broader cancer informatics landscape.

\begin{table*}[ht]
\centering
\small
\caption{A comparative analysis of prior research on cancer classification that is founded on the type of cancer being addressed, the nature of the dataset, and the application of advanced modeling techniques.  Each row denotes a study and specifies whether the dataset utilized is publicly accessible (\ding{51}), whether the study encompasses multiple cancer types (\ding{51}), and whether transformer-based models were implemented (\ding{51}).  The absence of that characteristic is denoted by a \ding{55}.  The majority of current methodologies are dependent on private or single-type datasets, which have restricted model generalization.  Conversely, the proposed R-GAT model emphasizes its accessibility, generalizability, and methodological advancement by incorporating transformer-based mechanisms, supporting multi-cancer classification (Thyroid, Colon, and Lung), and utilizing public biomedical abstracts.}
\label{table:cancer-classification}

\renewcommand{\arraystretch}{1.2}

\resizebox{1.05\textwidth}{!}{%
\begin{tabular}{l p{3.5cm} p{5.5cm} c c c}
\toprule
\textbf{Study} & \textbf{Cancer Type(s)} & \textbf{Data Type} & \textbf{Public?} & \textbf{Multi-cancer} & \textbf{Transformer} \\
\midrule
Nguyen et al. \cite{nguyen2020hybrid} & Breast Cancer & Radiology Reports & \ding{55} & \ding{55} & \ding{55} \\
Tang et al. \cite{Tang2019} & N/A & Clinical Progress Notes & \ding{55} & \ding{55} & \ding{51} \\
Alachram et al. \cite{alachram2021text} & Breast and Other Cancers & PubMed Abstracts, Gene Expression Data & \ding{55} & \ding{51} & \ding{55} \\
Uskaner et al. \cite{uskaner2023using} & Breast Cancer & Mammography Reports & \ding{55} & \ding{55} & \ding{51} \\
Jasmir et al. \cite{jasmir2020text} & N/A & Clinical Trial Documents & \ding{51} & \ding{55} & \ding{55} \\
Prabhakar et al. \cite{prabhakar2021medical} & N/A & Clinical Patient Records & \ding{55} & \ding{55} & \ding{55} \\
Achilonu et al. \cite{achilonu2021text} & Breast, Colorectal, Prostate & Pathology Reports (Free-text) & \ding{55} & \ding{51} & \ding{55} \\
Mithun et al. \cite{mithun2023clinical} & Lung Carcinoma & Radiology Reports & \ding{55} & \ding{55} & \ding{55} \\
Du et al. \cite{du2019ml} & N/A & Biomedical Literature, Clinical Notes & \ding{55} & \ding{55} & \ding{55} \\
\rowcolor{green!10}
\textbf{Proposed R-GAT Model} & Thyroid, Colon, Lung & Biomedical Abstracts & \ding{51} & \ding{51} & \ding{51} \\
\bottomrule
\end{tabular}%
} 

\end{table*}

\end{document}